\pdfoutput=1

\documentclass[11pt]{article} 

\usepackage[]{acl2023}

\usepackage{times}
\usepackage{latexsym}

\usepackage[T1]{fontenc}

\usepackage[utf8]{inputenc}

\usepackage{microtype}

\usepackage{inconsolata}

\usepackage{blindtext}

\usepackage{url}
\usepackage{latexsym}
\usepackage{numprint}

\usepackage{graphicx}
\usepackage{natbib}
\usepackage{amsmath}
\usepackage{latexsym}
\usepackage{booktabs}
\usepackage{color}
\usepackage{wrapfig}

\usepackage{epsfig}
\usepackage[ruled,vlined]{algorithm2e}

\usepackage{multirow}
\usepackage{multicol}
\usepackage[textsize=scriptsize]{todonotes}

\title{Discourse Analysis via Questions and Answers: \\ Parsing Dependency Structures of Questions Under Discussion}

\author{Wei-Jen Ko$^1$\ \ \ \
Yating Wu$^2$\ \ \ \
Cutter Dalton$^3$\ \ \ \
Dananjay Srinivas$^3$\\ 
\textbf{Greg Durrett}$^1$\ \ \ \
\textbf{Junyi Jessy Li}$^4$\\
$^1$ Computer Science, $^2$ Electrical and Computer Engineering, $^4$ Linguistics, \\
The University of Texas at Austin\\
$^3$ Linguistics, University of Colorado Boulder\\
{\small \tt wjko@outlook.com, yating.wu@utexas.edu, 
cutter.dalton@colorado.edu, 
dananjay.srinivas@gmail.com,}\\
{\small \tt gdurrett@cs.utexas.edu,
jessy@utexas.edu}\\
}

\begin{document}
\maketitle

\begin{abstract}
Automatic discourse processing
is bottlenecked by data: current discourse formalisms pose highly demanding annotation tasks involving large taxonomies of discourse relations, making them inaccessible to lay annotators. This work instead adopts the linguistic framework of Questions Under Discussion (QUD) for discourse analysis and seeks to derive QUD structures automatically. QUD views each sentence as an answer to a question triggered in prior context; thus, we characterize relationships between sentences as free-form questions, in contrast to exhaustive fine-grained taxonomies. We develop the first-of-its-kind QUD parser that derives a dependency structure of questions over full documents, trained using a large, crowdsourced question-answering dataset DCQA~\cite{dcqa}.
Human evaluation results show that QUD dependency parsing is possible for language models trained with this  crowdsourced, generalizable annotation scheme.
We illustrate how our QUD structure is distinct from RST trees, and demonstrate the utility of QUD analysis in the context of document simplification. Our findings show that QUD parsing is an appealing alternative for automatic discourse processing.
\end{abstract}

\section{Introduction} 

Discourse structure characterizes how each sentence in a text relates to others to reflect the author's high level reasoning and communicative intent.
Understanding discourse can be widely useful in applications such as text summarization~\cite{hirao2013single,gerani-etal-2014-abstractive,durrett2016learning,xu2020discourse}, classification~\cite{bhatia-etal-2015-better,ji2017neural}, narrative understanding~\cite{lee2019multi}, machine comprehension~\cite{narasimhan2015machine}, etc. 

However, automatically inferring discourse structure is challenging which
hinders wider application~\cite{atwell2021we}. At its root lies the issue of data annotation:
popular coherence formalisms like the Rhetorical Structure Theory (RST,~\citet{rst}, Segmented Discourse Representation Theory (SDRT,~\citet{asher2003logics}, and the Penn Discourse Treebank (PDTB,~\citet{pdtb} require
experts---typically linguists trained for the task---to reason through long documents over  large relation taxonomies. These features, coupled with the difficulties of annotating full structures in the case of RST and SDRT, make the task inaccessible to lay annotators. The taxonomies differ across formalisms~\cite{demberg2019compatible}, and their coverage and definitions are being actively researched and refined~\cite{sanders1992toward,taboada2006rhetorical,prasad2014reflections}.

\begin{figure*}[t]
\begin{center}
\includegraphics[width=0.95\textwidth]{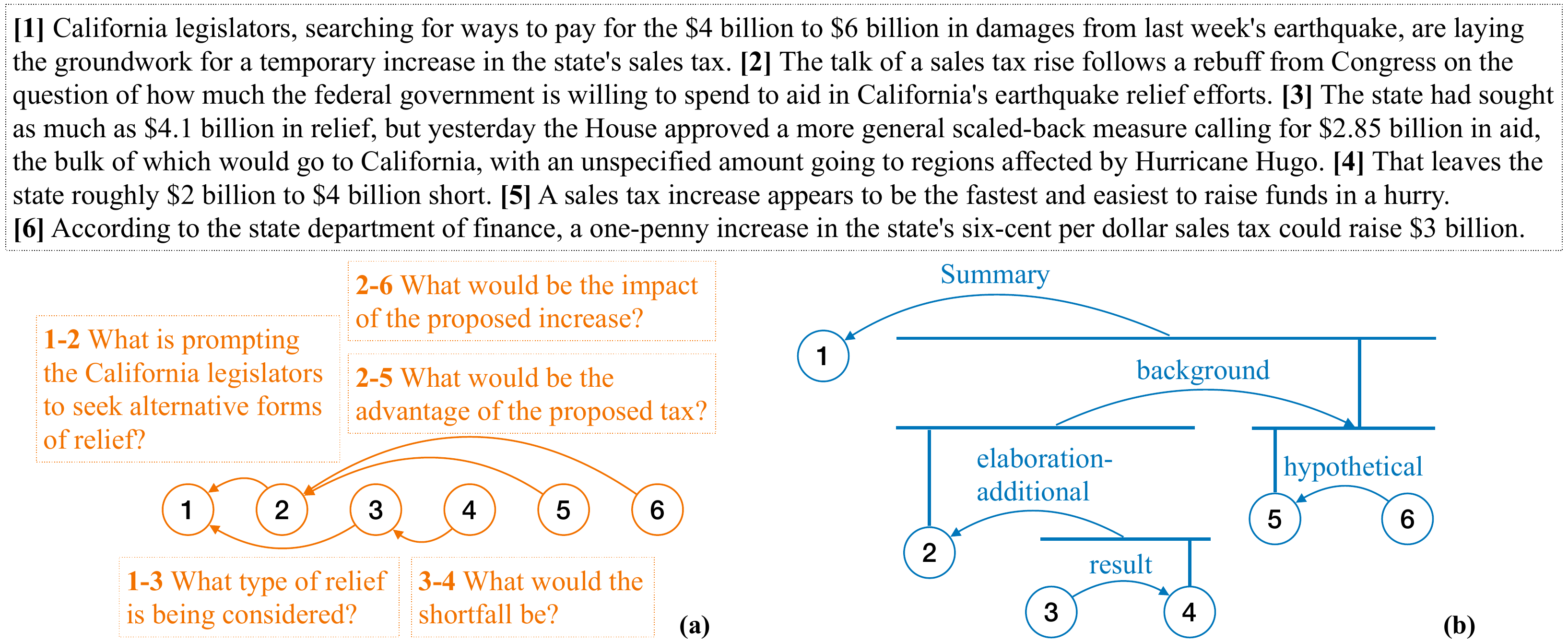}
\end{center}
\vspace{-0.5em}
\caption{A snippet of a WSJ article from the intersecting subset of DCQA~\cite{dcqa} and the RST Discourse Treebank~\cite{rstdt}. (a) shows a QUD dependency structure derived from DCQA. Edges are defined by questions, connecting where the question arose from (the ``anchor'' sentence) and the sentence that answers the question. (b) shows the annotated RST tree above the sentence level.}
\label{fig:tree-comparison}
\end{figure*}

In contrast, this work aims to derive discourse structures that fit into the linguistic framework of \emph{Questions Under Discussion} (QUD)~\cite{von1989referential,van1995discourse}, which neatly avoids reliance on a strict taxonomy.
In QUD, ``each sentence in discourse addresses a (often implicit) QUD either by answering it, or by bringing up another question that can help answering that QUD. The linguistic form and the interpretation of a sentence, in turn, may depend on the QUD it addresses''
~\cite{benz2017questions}. 
Thus relationships
between sentences can be characterized by free-form questions instead of pre-defined taxonomies. For instance, consider the following two sentences:
\advance\leftmargini -1em
\begin{quote}
\small
\textbf{(S3)}: A route out of Sarajevo was expected to open later today — but only for international humanitarian agencies that already can use another route.\\
\textbf{(S6)}: A four-month cease-fire agreement signed Dec. 31 made possible the medical evacuation and opening of the route into Sarajevo today.
\end{quote}
Sentence 6 is the answer to a question from sentence 3: \emph{``Why can they open a route?''}.
The question-answer view is in line with recent work reformulating linguistic annotation as question answering ~\cite{he2015question,pyatkin2020qadiscourse,klein2020qanom}, which reduces the bar for data collection and allows advancements in QA systems to be recruited~\cite{aralikatte2021ellipsis}. Furthermore, QUD's reliance on natural language annotation aligns with large language models (e.g., GPT-3) using language as a universal ``interface'' across various tasks.

Despite the richness in theoretical research related to QUD, data-driven efforts are scarce;
recent work has started corpora development under QUD~\cite{de2018qud,tedq,hesse2020annotating}, but these dedicated datasets are small and no computational models have yet been built to automatically derive QUD structures. 

This work seeks to fill this gap, and presents the first-of-its-kind QUD parser. This parser takes a document as input and returns a question-labeled dependency structure over the sentences in the document, as depicted in Figure~\ref{fig:tree-comparison}(a). 
For training, we use the intra-document question answering dataset DCQA~\cite{dcqa}; DCQA's annotation scheme is both compatible with QUD and easily crowdsourced, making QUD parsing a much less costly option than existing frameworks.

Each question in DCQA is considered to arise from an ``anchor'' sentence, and answered by another sentence later in the same article.
In line with QUD, we consider each sentence as the answer to an implicit question from prior context~\cite{hunter2015rhetorical}, in particular the anchor sentence. 
We view the anchor sentence as the parent node of the answer sentence, with the question describing the relation between the two; this results in a dependency tree structure.

Conveniently, a subset of DCQA overlaps with the RST Discourse Treebank~\cite{rstdt}, allowing us to directly compare the two types of structures (Figure~\ref{fig:tree-comparison}(b)).
We show that the QUD trees are structurally distinct from RST trees. A close inspection of relation-question correspondence reveals that QUD's free-form questions are more fine-grained, and that their presence reduces annotator disagreement in selecting RST relations.

Trained on DCQA, our QUD parser consists of two models used in a pipeline. The first model predicts the anchor sentence for each (answer) sentence in the article;
the second model performs question generation given the answer sentence and the predicted anchor sentence. 
Our comprehensive human evaluation shows that readers approve of 71.5\% of the questions generated by our best model; among those, the answer sentence answers the generated question 78.8\% of the time.
Finally, we demonstrate the analytical value of QUD analysis in the context of news document simplification: the questions reveal how content is elaborated and reorganized in simplified texts.

In sum, this work marks the first step in QUD parsing; our largely positive human evaluation results show that this is a promising data-driven approach to discourse analysis with \emph{open, crowdsourced} annotation that is so far infeasible to do at scale with other discourse frameworks.
We release our models at \url{https://github.com/lingchensanwen/DCQA-QUD-parsing}.

\section{Background and related work}\label{sec:related}
\paragraph{Discourse frameworks}
Questions Under Discussion is a general framework 
with vast theoretical research especially in pragmatics, e.g., information structure~\cite{roberts2012information,buring2003d,velleman2016question}, presuppositions~\cite{simons2010projects}, and  implicature~\cite{hirschberg1985theory,van1996inferring,jasinskaja2017discourse}.
\citet{ginzburg1996dynamics} extended \citet{stalnaker1978assertion}'s dynamic view of context to dialogue by integrating QUD with dialogue semantics, where the speakers are viewed as interactively posing and resolving queries.
In QUD analysis of monologue, each sentence aims to answer a (mostly implicit) question triggered in prior context. 
Sometimes the questions form hierarchical relationships (stacks where larger questions have sub-questions, starting from the root question ``\emph{What is the way things are?}'')~\cite{buring2003d,roberts2004context,de2018qud,riester2019constructing}. However, because of the inherent subjectivity among naturally elicited QUD questions~\cite{tedq,inq}, we leave question relationships for future work.

QUD and coherence structures 
are closely related. Prior theoretical work looked into the mapping of QUDs to discourse relations~\cite{jasinskaja2008explaining,onea2016potential} or the integration of the two~\cite{kuppevelt1996directionality}. \citet{hunter2015rhetorical} and \citet{riester2019constructing} studied structural correspondances between QUD stacks and SDRT specifically. \citet{tedq} 
showed that QUD could be a useful tool to quantitatively study the predictability of discourse relations~\cite{garvey1974implicit,kehler2008coherence,bott2014verbs}. In~\citet{pyatkin2020qadiscourse}, discourse relation taxonomies were also converted to templatic questions, though not in the QUD context.

Traditionally, discourse ``dependency parsing'' refers to parsing the RST structure~\cite{hirao2013single,bhatia-etal-2015-better,morey2018dependency}.
Since QUD structures are marked by free-form questions, the key aspect of ``parsing'' a QUD structure is thus question generation,
yielding a very different task and type of structure than RST parsing. As we show in the paper, the two are complementary to each other and not comparable. This work focuses on automating and evaluating a QUD parser; we leave for future work to explore what types of structure is helpful in different downstream tasks.

\vspace{-0.3em}
\paragraph{The DCQA dataset}
Corpora specific for QUD are scarce. Existing work includes a handful of interviews and 40 German driving reports annotated with question stacks~\cite{de2018qud,hesse2020annotating}, as well as \citet{tedq}'s 6 TED talks annotated following~\citet{kehler2017evaluating}'s expectation-driven model (eliciting questions without seeing upcoming context). \citet{inq}'s larger {\sc inquisitive} question dataset is annotated in a similar manner, but {\sc inquisitive} only provides questions for the first 5 sentences of an article, and they did not annotate answers.

This work in contrast repurposes the much larger \textbf{DCQA} dataset~\cite{dcqa}, 
consisting of more than 22K questions crowdsourced across 606 news articles. 
DCQA was proposed as a way to more reliably and efficiently collect data to train QA systems to answer high-level questions, specifically QUD questions in {\sc inquisitive}. Though not originally designed for QUD parsing, DCQA is suitable for our work because its annotation procedure follows the reactive model of processing that is standard in QUD analysis~\cite{benz2017questions}, where the questions are elicited after observing the upcoming context. 
Concretely, for each sentence in the article, the annotator writes a QUD such that the sentence is its answer, and identifies the ``anchor'' sentence in preceding context that the question arose from. Figure~\ref{fig:tree-comparison}(a) shows questions asked when each of the sentences 2-6 are considered as answers, and their corresponding anchor sentences. As with other discourse parsers, ours is inevitably bound by its training data. However, DCQA's crowdsourcable paradigm makes future training much easier to scale up and generalize.

\section{Questions vs.\ coherence relations}\label{sec:rstrelations}

We first illustrate how questions capture inter-sentential relationships, compared with those in coherence structures. We utilize the relation \emph{taxonomy} in RST for convenience, as in Section~\ref{sec:treestats} we also compare the structure of our QUD dependency trees with that of RST.

Given each existing anchor-answer sentence pair across 7 DCQA documents, we asked two graduate students in Linguistics to select the most appropriate discourse relation between them (from the RST relation taxonomy~\cite{carlson2001discourse}). Both students were first trained on the taxonomy using the RST annotation manual.

\vspace{-0.3em}
\paragraph{Analysis}

The frequency distribution of annotated RST relations that occurred $\geq10$ times (counting each annotator independently) is:
\emph{elaboration(200), cause(75), manner-means(69), background(64), explanation(55), comparison(33), condition(32), contrast(17), temporal(15), attribution(14)}. E.g.,
\begin{quote}
\small
    \textbf{\emph{[context]}} Early one Saturday in August 1992, South Floridians discovered they had 48 hours to brace for, or flee, ... one of the nation's most infamous hurricanes. \textbf{\emph{[anchor]}} Oklahomans got all of 16 minutes before Monday's tornado.
    \\
    \textbf{\emph{[QUD]}} How much time do people normally have to prepare for tornadoes?
    \\
    \textbf{\emph{[answer]}} And that was more time than most past twisters have allowed.
    
    \textbf{RST label:} Comparison
\end{quote}

Our analysis shows that the questions are often more fine-grained than RST relation labels; in the example below, the QUD describes what is being elaborated:
\begin{quote}
\small
\textbf{\emph{[anchor]}} Crippled in space, the Kepler spacecraft’s planet-hunting days are likely over.
    \\
    \textbf{\emph{[QUD]}} What plans does NASA have for the damaged spacecraft?
    \\
    \textbf{\emph{[answer]}} Engineers will try to bring the failed devices back into service, or find other ways to salvage the spacecraft.

    \textbf{RST label:} Elaboration-Additional
\end{quote}


Agreeing on what is the most appropriate RST relation, as expected, is difficult with its large relation taxonomy: Krippendorff's $\alpha$ (with MASI distance to account for multiple selection) between the two annotators is 0.216, indicating only fair agreement~\cite{artstein2008inter}. To study the effects of seeing the QUD, we further asked the annotators to find a relation \emph{without} the question.\footnote{We paced 3 months between annotation with and without the question to minimize memorization effects.} This led to a much lower, 0.158 $\alpha$ value.
Thus the presence of the QUD could, in some cases, align divergent opinions, as in the following example:
\begin{quote}
\small
    \textbf{\emph{[context]}} For the past four years, the \$600 million Kepler has been a prolific planet detector from its lonely orbit... \textbf{\emph{[anchor]}} The project has been a stunning success, changing our view of the universe.
    \\
    \textbf{\emph{[QUD]}} What knowledge did we have about solar systems before the project?
    \\
    \textbf{\emph{[answer]}} Before Kepler, we knew little about other solar systems in the Milky Way galaxy.

    \textbf{RST labels with questions:} Background; Background
    \\
    \textbf{RST labels w/o questions:} Evidence; Circumstance
\end{quote}

We also find that sometimes a question could be interpreted in terms of different RST relations:
\begin{quote}
\small
 \textbf{\emph{[anchor]}} According to a preliminary National Weather Service summary, Monday's tornado was a top-end EF5, with top winds of 200 to 210 miles per hour (mph), and was 1.3 miles wide.
    \\
    \textbf{\emph{[QUD]}} How long did the tornado last?
    \\
    \textbf{\emph{[answer]}} It was tracked on the ground for 50 minutes - an eternity for a tornado - and its damage zone is more than 17 miles wide.
    
    \textbf{RST labels that could work:} Evidence, Proportion, Elaboration-Additional, Manner
\end{quote}

These findings indicate that while questions often relate to coherence relations, they are typically more specific and can also capture aspects from multiple relations. This  supports \citet{hunter2015rhetorical}'s skepticism about the correspondence of QUD and coherence structures, though they focused more on structural aspects of SDRT.

\section{Deriving QUD dependency structures}

Our task is to derive a QUD dependency structure over a document $D = (\mathbf{s}_1,\ldots,\mathbf{s}_n)$ consisting of $n$ sentences. A QUD tree $\mathbf{T} = ((a_1,\mathbf{q}_1),\ldots,(a_n,\mathbf{q_n}))$ can be expressed as a list of $n$ tuples: each sentence has an associated anchor sentence $a_i$ and a question labeling the edge to the anchor $\mathbf{q}_i$. To arrive at a dependency structure, we view the anchor sentence as the head of an edge, linking to the answer sentence via the question, as shown in Figure~\ref{fig:tree-comparison}(a).

We set $a_1 = 0$ and $\mathbf{q}_1 = \emptyset$; the first sentence is always the root of the QUD dependency tree, so has no parent and no question labeling the edge. Each other $a_i \in \{1,2,\ldots,i-1\}$ and $\mathbf{q}_i \in \Sigma^*$ for a vocabulary $\Sigma$.
We note that $\mathbf{T}$ is analogous to a labeled dependency parse, except with questions $\mathbf{q}$ in place of typical discrete edge labels.
Our parser is a discriminative model
\vspace{-0.5em}
\begin{equation}
\notag
    P(\mathbf{T} \mid D) = \prod_{i=1}^n \left[P_a(a_i \mid D, i) P_q(\mathbf{q}_i \mid D, i, a_i)\right].
\vspace{-0.5em}
\end{equation}
This formulation relies on models corresponding to two distinct subtasks. First, \emph{anchor prediction} selects the most appropriate sentence in prior context to be the anchor sentence of the generated question using a model $P(a_i \mid D, i)$.
Second, \emph{question generation} given the current (answer) sentence, its anchor, and the document context uses a model $P(\mathbf{q}_i \mid D, i, a_i)$.

We do not impose projectivity constraints or other structural constraints beyond anchors needing to occur before their children. Therefore, inference can proceed with independent prediction for each sentence.\footnote{We make a further simplifying assumption by doing greedy prediction of each $a_i$ before generating $\mathbf{q}$. We sample $\mathbf{q}$ using nucleus sampling and do not rely on the question probabilities to be informative about whether the structure itself is well-formed.} We now proceed to describe the models for $P_q$ and $P_a$ that constitute the parser.

\begin{figure}[t]
\begin{center}
\includegraphics[width=0.49\textwidth,trim=0mm 100mm 85mm 30mm]{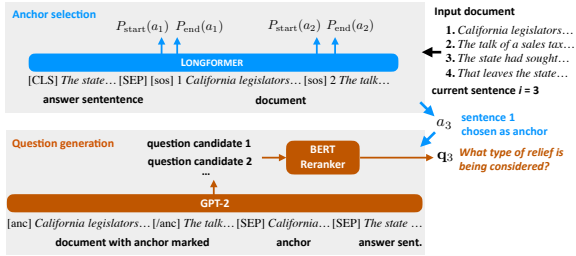}
\end{center}
\vspace{-0.5em}
\caption{Breakdown of the two modules used in the parser: anchor selection using a Longformer QA model to select an anchor index and question generation conditioned on the selected anchor and the answer, including a BERT reranking phase.}
\vspace{-0.5em}
\label{fig:dcqa-parser}
\end{figure}

\subsection{Anchor prediction}
The anchor prediction model $P_a$ considers the given sentence $\mathbf{s}_i$ and reasons through prior article context to find the most likely sentence where a QUD can be generated, such that $\mathbf{s}_i$ is the answer. Since this 
task involves long document contexts,
we use the Longformer model (\texttt{longformer-base-4096})~\cite{lfm}, 
shown to improve both time efficiency and performance on a range of tasks with long contexts.

We adopt the standard setup of BERT for question answering \cite{BERT} and model $P(a_i)$ as a product of start and end distributions.
For the input, we concatenate the answer sentence 
and the article as a single long sequence, separated by delimiters: \texttt{[CLS] [answer sentence] [SEP] [document]}. Following \citet{dcqa}, we add two tokens: the start of sentence token \texttt{[sos]} and the sentence ID, before every sentence in the article. We train the model to predict the span of the two added tokens in front of the anchor sentence.

We modify the HuggingFace~\cite{wolf-etal-2020-transformers} codebase for our experiments. We use the Adam \cite{adam} optimizer with $(\beta_1,\beta_2)=(0.9,0.999)$ and learning rate 5e-5. The model is trained for 25000 steps using batch size 4. We use the same article split for training, validation and testing as in DCQA, and the parameters are tuned on the validation set.

\subsection{Question generation}
Our question generator $P_q(\mathbf{q}_i \mid D, i, a_i)$ takes in the answer sentence $\mathbf{s}_i$ indexed by $i$, the anchor sentence at $a_i$, and the article $D$, and aims to generate an appropriate QUD. 
We fine-tune GPT-2 \citep{gpt2} for this purpose; \citet{inq} showed that GPT-2 generates open-ended, high-level questions with good quality.
To fine-tune this model, each input instance is a concatenation of four parts, separated by delimiters: (1) $\mathbf{s}_0,\mathbf{s}_1,...,\mathbf{s}_{i-1}$,
with the start and end of the anchor sentence marked by special tokens; (2) the anchor sentence; (3) $\mathbf{s}_i$; (4) the question. 

\vspace{-0.3em}
\paragraph{Inference} During inference, we feed in (1)---(3) and sample from the model to generate the question. By default, we use nucleus sampling \citep{nucleus} with $p=0.9$. To improve the consistency of questions with the anchor or answer sentences, we use an additional \textbf{reranking step}.

Our reranker is a BERT binary classification model formatted as \texttt{[CLS] [question] [SEP] [anchor sentence] [answer sentence]}.
Positive examples consist of annotated questions, anchor, and answer sentences in the DCQA training set; we synthetically generate negative examples by replacing the anchor or answer sentences with others in the same article. Training is detailed in Appendix~\ref{app:reranker}.
To rerank, we sample 10 questions from the generation model, and choose the question with the highest posterior from the reranker. 

\vspace{-0.3em}
\paragraph{Reducing question specificity} We found that questions generated by the above model often copy parts of the answer sentence, including information that is introduced for the first time in the answer sentence. For example, in Figure~\ref{fig:tree-comparison}, Hurricane Hugo is first mentioned in sentence 3. The model might ask ``\emph{What type of relief is going to California and regions affected by Hurricane Hugo?}'' This makes the question prone to ``foresee'' details that are unlikely to be inferred from previous context, violating QUD principles. 
We observe that these unwanted details often pertain to specific entities. To this end, in the answer sentence, we
replace each token that belongs to a named entity with its entity type before feeding into the GPT-2 model.\footnote{We use the \texttt{bert-base-NER} model trained on the CoNLL-2003 NER dataset \cite{tjong-kim-sang-de-meulder-2003-introduction}}

\begin{figure}[t]
\begin{center}
\includegraphics[width=0.45\textwidth]{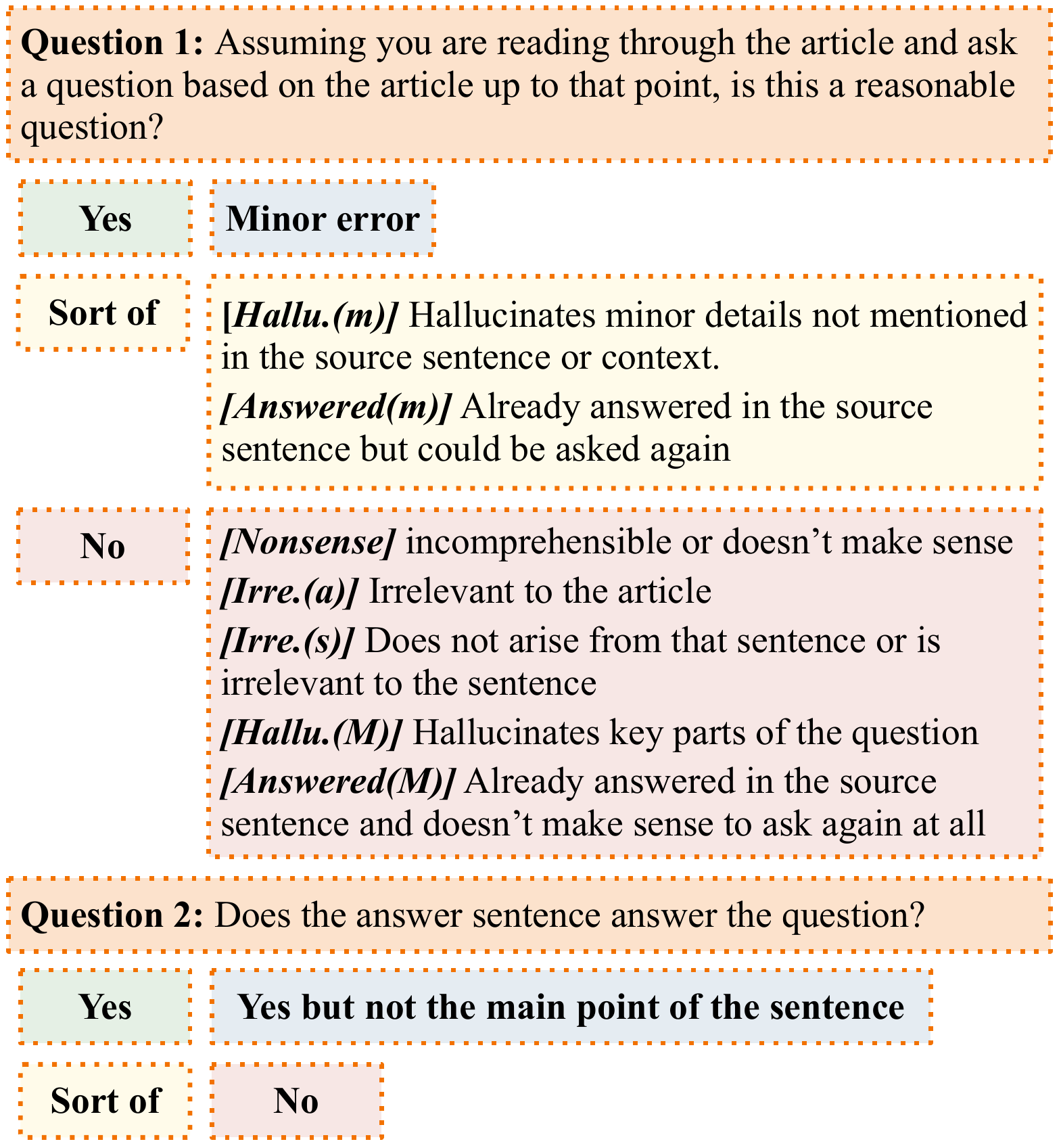}
\end{center}
\vspace{-1em}
\caption{Evaluation schema.}
\vspace{-0.5em}
\label{fig:eval-schema}
\end{figure}

\section{Evaluation and analysis}\label{sec:eval}
Since QUD parsing features an open-ended generation component, we need new evaluation methodology compared to standard discourse parsing. We focus on two main factors: (1) whether the generated question is plausible at the predicted anchor point; (2) whether the question is actually answered by the answer sentence.

In QUD annotation and DCQA itself~\cite{tedq,inq,dcqa}, it is often the case that multiple questions can be asked even given the same anchor and/or answer sentences.
The evaluation of QUD validity thus involves complex reasoning performed jointly among (long) context, the anchor, the answer sentence, and the generated question itself. For these reasons, we rely on human evaluation, and leave the development of automatic evaluation metrics for future work.\footnote{Existing automatic measures for open-ended tasks are known to correlate poorly with human judgments~\cite{howcroft2020twenty,celikyilmaz2020evaluation}; additionally, whether the answer sentence actually answers the question is a key aspect to validate QUD. But as shown in~\citet{dcqa}, poor performance bars existing QA models from being used to evaluate QUD parsing. Appendix~\ref{app:anchor_prediction} discusses anchor prediction ``accuracies'' against human-annotated anchors in DCQA.}

\subsection{Human Evaluation Setup}\label{sec:eval-setup}

\begin{table*}
\small
\centering
\begin{tabular}{l|c|c|cc|ccccc}
\toprule
      \textbf{System} &\textbf{Yes}&\textbf{Minor error}& \multicolumn{2}{c|}{\textbf{Sort of}}&\multicolumn{5}{c}{\textbf{No}}  \\ 
      &&&Hallu.(m)&Ans.(m)&Nonsense&Irre.(a)&Irre.(s)&Hallu.(M)&Ans.(M)  \\ \midrule
Full &71.5&4.2&7.1&4.0&6.4&0.2&3.0&2.4&1.2     \\
~~~~-Reranking &66.7&3.4&8.4&4.5&6.3&0.2&7.8&1.7&1.0\\
~~~~~~~~-NER&   54.8&2.8&10.7&4.2&6.2&0.6&16.9&2.9&1.0   \\
\bottomrule
\end{tabular}
\vspace{-0.5em}
\caption{Human evaluation results for Question 1.}
\vspace{-0.5em}
\label{tab:human1}
\end{table*}

\begin{table}
\centering
\small
\begin{tabular}{l|cccc}
\toprule
      \textbf{System} &\textbf{Yes}&\textbf{Not main point}& \textbf{Sort of} &\textbf{No} \\  \midrule
Full & 78.8&3.1&10.5&7.6  \\
~~~~-Reranking &71.8&1.8&14.1&12.3\\
~~~~~~~~-NER&    76.7&2.8&11.0&9.4\\
\bottomrule

\end{tabular}
\vspace{-0.5em}
\caption{Human evaluation results for Question 2.}
\label{tab:human2}
\end{table}

Our evaluation task shows human judges the full article, the anchor and answer sentences, and the generated question. We then ask them to judge the quality of the generated QUD
using a hierarchical schema shown in Figure~\ref{fig:eval-schema}. The criteria in our evaluation overlap with~\citet{de2018qud}'s human annotation guidelines, while specifically accommodating typical errors observed from machine-generated outputs.

\textbf{Question 1 (Q1)} assesses how reasonable the question is given context prior to and including the anchor sentence. The judges have four graded options: (1) \emph{yes} for perfectly fine questions; (2) \emph{minor error} for questions that contain minor typos or grammatical errors that do not impact its overall good quality; (3) \emph{sort of} for questions with non-negligible though not catastrophic errors; and (4) \emph{no} for questions that are not acceptable. (3) and (4) both contain subcategories representative for a sample of questions we closely inspected a priori.

\textbf{Question 2 (Q2)} assesses whether the question is answered by the targeted answer sentence, also with four graded options: (1) \emph{yes} where the targeted answer sentence is clearly an answer to the generated question; (2) \emph{yes but not the main point} where 
the answer is not the at-issue content of the answer sentence. Such cases violate Grice's principle of quantity~\cite{grice1975logic} and QUD's principle that answers should be the at-issue content of the sentence~\cite{simons2010projects}. (3) \emph{sort of} where the answer sentence is  relevant to the question but it is questionable whether it actually addresses it; and (4) \emph{no} where the generated question is clearly not addressed by the answer sentence. Annotators are allowed to skip Q2 if the generated question from Q1 is of lower quality.

\subsection{Results}

We recruited 3 workers from Mechanical Turk as judges who have an established relationship with our lab, and are experienced with tasks involving long documents. 
They are compensated above \$10 per hour. We annotate 380 questions from 20 articles from the DCQA test set.
Inter-annotator agreement is reported in Appendix~\ref{app:agreement}.

\paragraph{Q1 results} As seen in Table~\ref{tab:human1}, for our full model, 71.5\% of responses are ``yes''es, showing that most of the generated questions are of good quality. Without reranking, there are 4.8\% fewer ``yes'' responses; there are more questions that do not rise from the anchor sentence, showing the effectiveness of our reranker.
Further removing NER masking results in a substantial drop of 11.9\% of good questions. There are also more questions hallucinating details and/or irrelevant to the anchor sentence.

\paragraph{Q2 results} Since Question 2 may not make sense when the generated question is of low quality, we show the results of Q2 on a subset of questions where all three workers answered ``yes'' or ``minor error'' for Q1 (see Table \ref{tab:human2}).  
Of those questions, annotators chose ``yes'' 78.8\% of the time,
showing that a majority of good-quality questions are actually answered in the answer sentence and represents anchor-answer sentence relationships. Our full model has better performance than the two ablations, showing the effectivenes of reranking. Further, since masking NER removes some of the information from the answer sentence, the percentage of ``yes''es is slightly lower after masking.

\begin{figure}[t]
    \centering
    \includegraphics[width=0.45\textwidth]{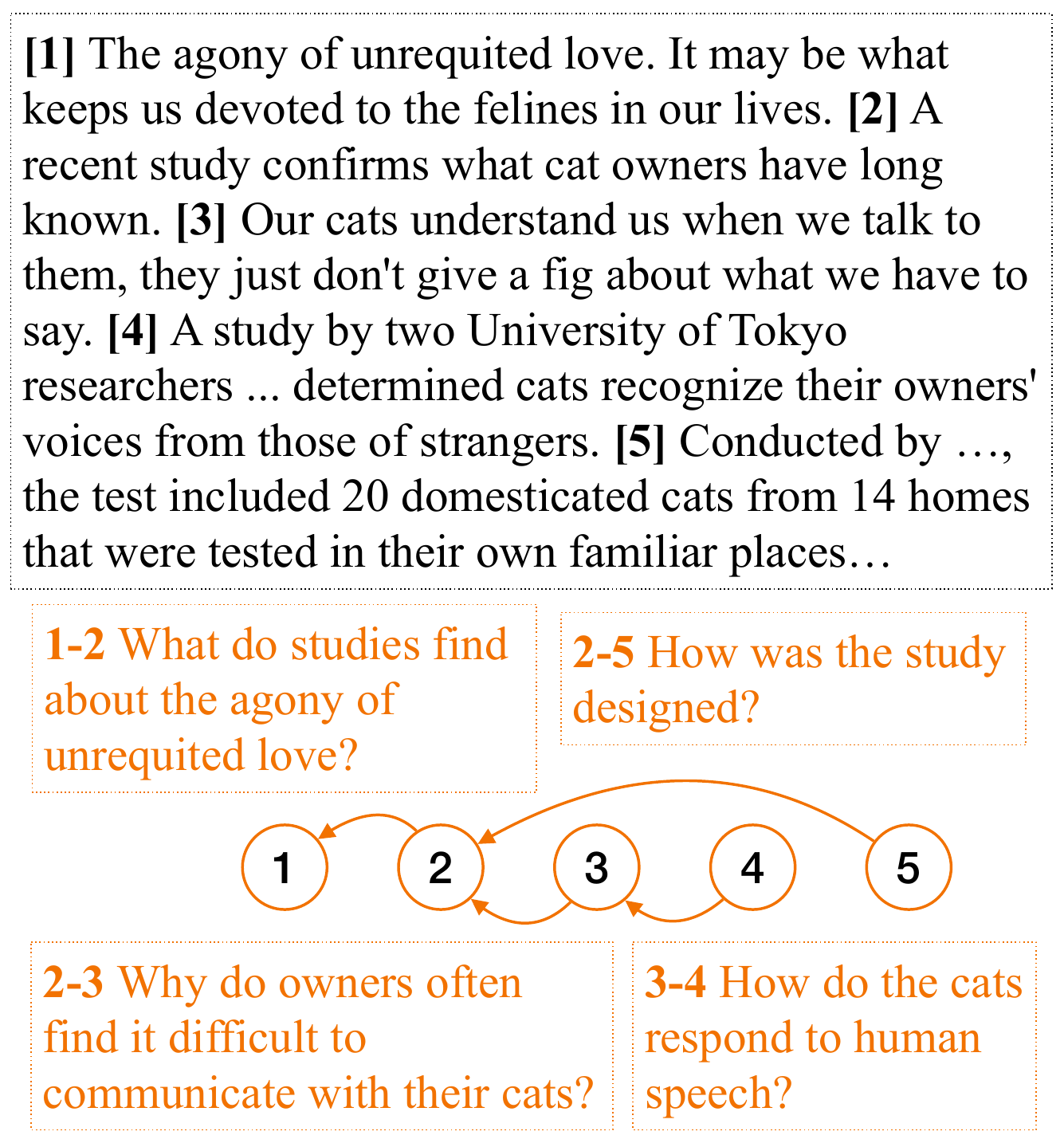}
    \vspace{-0.5em}
    \caption{Example of model-generated QUD structure.}
    \vspace{-0.5em}
    \label{fig:output1}
\end{figure}

These results show that most of the time, our full system is able to generate questions that are good in terms of linguistic form and are also reasonable QUD questions given prior context. Most of these good questions are clearly answered in the answer sentence, i.e., they legit questions under the reactive model of mental processing. These results indicate a strong QUD parser with a large portion of valid QUD links.
In Figure~\ref{fig:output1} and Appendix~\ref{app:outputs}, we visualize  output examples.

\begin{table*}
\centering
\small
\begin{tabular}{l|l|ccccc|c}
\toprule
      \textbf{data} &\textbf{tree type}     & \textbf{height}& \textbf{norm.\ arc len.}&\textbf{prop.\ of leaf}&\textbf{avg.\ depth}&\textbf{right branch}&\textbf{att.\ score} \\ \midrule
RST $\cap$ DCQA& RST-dep   &5.86&0.12&0.53&3.49&0.40&\multirow{2}{*}{0.30}         \\
RST $\cap$ DCQA& DCQA-human  & 6.72&0.21&0.48&3.88&0.45&        \\
\midrule
DCQA (val)& DCQA-human & 6.04&0.29&0.50&3.57&0.39& \multirow{2}{*}{0.47}          \\
DCQA (val)& DCQA-model  &6.76&0.22&0.43&3.85&0.52&        \\
\bottomrule

\end{tabular}
\vspace{-0.5em}
\caption{Statistics of discourse dependency trees, on the intersecting documents of RST-DT and DCQA (upper portion) and the DCQA validation set (lower portion).}
\label{tab:statistics}
\end{table*}

\subsection{Characterizing tree structures}\label{sec:treestats} 

We further characterize annotated and parsed QUD trees;
we also contrast QUD trees with RST, using the intersection of DCQA and RST-DT~\cite{rstdt}. 
We follow \citet{hirao2013single} to convert RST constituency trees to dependency trees using nuclearity information.
Since the leaves of QUD trees are sentences, we also treat sentences as the smallest discourse units for RST. 

We report the six metrics following \citep{ferracane-etal-2019-evaluating}: 
1) tree \textbf{height};
2) \textbf{normalized arc length}: the average number of sentences between edges, divided by the number of sentence $n$ in the article; 
3) \textbf{proportion of leaf nodes}: the number of leaf nodes divided by $n$;
4) \textbf{average depth} of every node in the tree;
5) \textbf{right branch}: the number of nodes whose parent is the immediate preceding sentence in the article, divided by $n$;
6) \textbf{attachment score}: count of sentences whose parent node is the same sentence among the two types of trees, divided by $n$,
the total number of sentences. 
This captures the similarity of the two types of trees.

Compared with annotated QUD trees, machine generated ones are slightly deeper and more right-branching (Table \ref{tab:statistics}). 
The normalized arc lengths indicate that our model is not merely finding the immediately preceding sentence as the anchor, although human annotated trees tend to have slightly longer arc lengths.
Machine-derived trees have a lower gap degree~\cite{yadav-etal-2019-formal} (13.2 on average on the validation set), compared to annotated ones (15.1 on average).

\subsection{QUD vs.\ RST}
Compared with RST (Table \ref{tab:statistics}), QUD trees have longer arc lengths, showing that
they more frequently represent relations between more distant sentence pairs. The tree height and average node depth of DCQA trees are larger than those of RST.

While nuclearity in RST is able to provide a hierarchical view of the text that has been used in NLP tasks, it comes with a highly contested~\cite{wolf2005representing,taboada2006rhetorical} strong compositionality assumption that ``whenever two large text spans are connected through a rhetorical relation, that rhetorical relation holds also between the most important parts of the constituent spans''~\cite{marcu1996building}. \citet{marcu1998build} showed that this assumption renders the derived structure alone insufficient in text summarization. In contrast, the QUD framework does not make such an assumption since it does not have the RST notion of nuclearity. During left-to-right reading, QUD  describes how each sentence resolves an implicit question posed in prior context, so QUD dependencies derived in this work are always rooted in the first sentence and ``parentage'' does not necessarily entail salience. Combined with observations from Section~\ref{sec:rstrelations}, we conclude that RST and QUD are complementary frameworks capturing different types of structure.

\begin{figure*}[t]
\begin{center}
\includegraphics[width=1\textwidth]{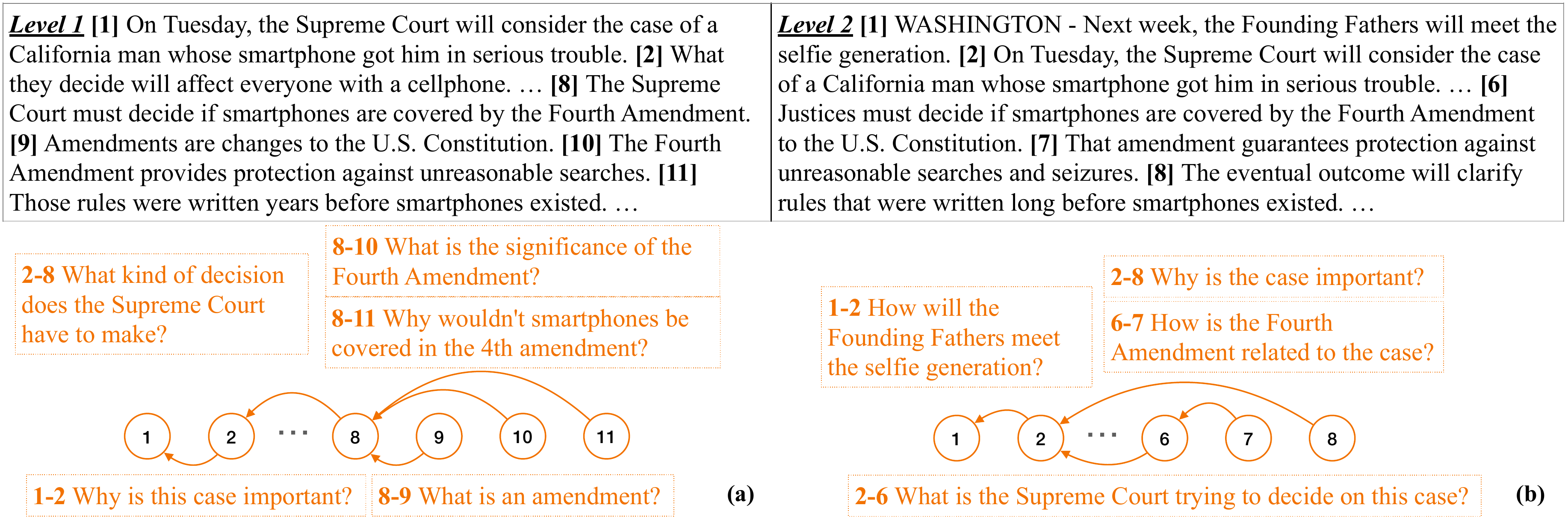}
\end{center}
\vspace{-0.5em}
\caption{Snippet of a Newsela article set with two reading levels.}
\label{fig:simplification}
\end{figure*}

\section{Case study: document simplification}\label{sec:simplification}

We demonstrate the analytical value of QUD analysis in the context of document simplification. We use the Newsela dataset~\cite{newsela}, where news articles are professionally simplified across several grade levels; a subset of Newsela (of the highest reading level) is present in DCQA. Note that most research in text simplification focus on the sentence level~\cite{alva2020data}; we hope to inform document-level approaches.

We sample 6 articles from the DCQA Newsela subset. For each of these, 3 linguistics undergraduates (not authors of this paper) doubly annotated their corresponding middle and elementary school levels
with QUD structures for the first 20 sentences following DCQA's paradigm. This amounts to $\sim$720 questions in total. Figure~\ref{fig:simplification} shows a snippet of our analysis from two reading levels of the same article.

We run and evaluate our parser on the articles of the second reading level. Using the schema in Figure~\ref{fig:eval-schema}, Question 1 is \emph{yes} for 60.2\% of the time, and Question 2 is \emph{yes} 75.2\% of the time. This shows that while the parser is still capable of generating reasonable questions, the performance degrades compared to testing on the highest level. This is likely due to clear stylistic, organizational, and vocabulary difference for simplified texts; for this reason, we resort to using annotated QUDs to illustrate idealized results for this analysis.

\paragraph{Analysis} The simplified articles, which mostly align with the original versions at the beginning, tend to contain significant reorganization of content especially later in the text. Nonetheless, we found that 62.2\% of the questions had a similar question on another reading level, reflecting that QUDs frequently stay invariant despite these differences.
For example, in Figure~\ref{fig:simplification}, the content of sentence 8 (level 2) is covered in sentence 2 (level 1), yet in both cases the question ``\emph{Why is the case important}'' is used to link these sentences. Similarly, questions \texttt{q2-6} (level 2) and \texttt{q2-8} (level 1), as well as questions \texttt{q6-7} (level 2) and \texttt{q8-10} (level 1) reflect the same QUD.

Often, articles from higher reading levels presupposes certain knowledge that gets \textbf{elaborated} or explained during simplification~\cite{srikanth2021elaborative}. QUD analysis informs how content should be elaborated: in Figure~\ref{fig:simplification}(a), the level 1 article defined the concept of amendment (question \texttt{q8-9}), absent in level 2. 

\textbf{Sentence splitting} as a frequent operation~\cite{petersen2007text,zhu2010monolingual,alva2020data} could also be explained by questions, as in the case of \texttt{q8-11} in level 1, which provides a rationale as to why sentence 8 in level 2 is split (into sentences 2 and 11 in level 1). Note that this explanation is rooted \emph{outside} of content conveyed by the sentence that was split.

Finally, editors also \textbf{omit} difficult content~\cite{petersen2007text,zhong2020discourse}, as in Figure~\ref{fig:simplification}(b): sentence 1 in level 2 is not present in the level 1 simplification (due to less salience and the reference to the ``selfie generation'' which goes beyond the targeted reading level). Level 2 thus contains the extra QUD: \texttt{q1-2}.

In sum, QUD analysis reveals how elaborated or omitted content fit into the larger context during simplification, potentially aiding future document-level simplification systems by providing intermediate rationales.

\section{Conclusion}
This work presents the first QUD (Questions Under Discussion) parser for discourse analysis. We derive dependency structures of QUD, viewing each sentence as an answer to a QUD triggered in an anchor sentence in prior context. This paradigm avoids costly annotation of coherence structures; rather, our parser can be trained on the crowdsourced dataset DCQA. We show strong parser performance with comprehensive human evaluation. We further demonstrate the richness of QUD analysis in document simplification.

\section{Limitations}
While our work is consistent with the key aspects of Questions Under Discussion, we do not attempt to take into account all aspects of this broad framework. Most notably, we do not model relationship between questions (or question stacks), as mentioned in Section~\ref{sec:related}. While such relationships are potentially useful, with question stacks, the annotation task becomes much more expensive; currently, no existing dataset is available to train parsers in this fashion. We applaud the development of tools such as TreeAnno~\cite{de2018qud} to aid annotation. Additionally, because questions are open-ended, they are inherently subjective, which adds substantial challenge to modeling and evaluating stacks. Constrained by DCQA's setup, we also do not explicitly model QUD with multi-sentence answers, and leave this for future work.

The subjectivity of QUD analysis also means that there is no single ``right'' structure. This is in contrast to coherence structures that more rigorously define their structures and relation taxonomies (multiple analyses still exist in those structures, but to a lesser degree). Nonetheless, we showed in Section~\ref{sec:simplification} that consistency is still present despite documents being reworded and restructured during simplification.

To evaluate our parser, we developed a human evaluation scheme. As mentioned in Section~\ref{sec:eval}, automatic evaluation of QUD structure contains both a generation and a question-answering component. However, human evaluation is costly; future work looking into the development of automatic evaluation measures can be extremely valuable.

\section*{Acknowledgments}
We thank Kathryn Kazanas, Keziah Reina, and Anna Alvis for their contributions on text simplification analysis. We thank David Beaver for helpful discussions and comments.
This research is partially supported by NSF grants IIS-2145479, IIS-2107524. We acknowledge the Texas Advanced Computing Center (TACC)\footnote{\url{https://www.tacc.utexas.edu}} at UT Austin for many of the results within this paper.

\bibliography{anthology,custom}

\begin{thebibliography}{72}
\expandafter\ifx\csname natexlab\endcsname\relax\def\natexlab#1{#1}\fi

\bibitem[{Alva-Manchego et~al.(2020)Alva-Manchego, Scarton, and
  Specia}]{alva2020data}
Fernando Alva-Manchego, Carolina Scarton, and Lucia Specia. 2020.
\newblock Data-driven sentence simplification: Survey and benchmark.
\newblock \emph{Computational Linguistics}, 46(1):135--187.

\bibitem[{Aralikatte et~al.(2021)Aralikatte, Lamm, Hardt, and
  S{\o}gaard}]{aralikatte2021ellipsis}
Rahul Aralikatte, Matthew Lamm, Daniel Hardt, and Anders S{\o}gaard. 2021.
\newblock Ellipsis resolution as question answering: An evaluation.
\newblock In \emph{Proceedings of the 16th Conference of the European Chapter
  of the Association for Computational Linguistics: Main Volume}, pages
  810--817.

\bibitem[{Artstein and Poesio(2008)}]{artstein2008inter}
Ron Artstein and Massimo Poesio. 2008.
\newblock Inter-coder agreement for computational linguistics.
\newblock \emph{Computational linguistics}, 34(4):555--596.

\bibitem[{Asher et~al.(2003)Asher, Asher, and Lascarides}]{asher2003logics}
Nicholas Asher, Nicholas~Michael Asher, and Alex Lascarides. 2003.
\newblock \emph{Logics of conversation}.
\newblock Cambridge University Press.

\bibitem[{Atwell et~al.(2021)Atwell, Li, and Alikhani}]{atwell2021we}
Katherine Atwell, Junyi~Jessy Li, and Malihe Alikhani. 2021.
\newblock Where are we in discourse relation recognition?
\newblock In \emph{Proceedings of the 22nd Annual Meeting of the Special
  Interest Group on Discourse and Dialogue}, pages 314--325.

\bibitem[{Beltagy et~al.(2020)Beltagy, Peters, and Cohan}]{lfm}
Iz~Beltagy, Matthew~E. Peters, and Arman Cohan. 2020.
\newblock Longformer: The long-document transformer.
\newblock \emph{arXiv:2004.05150}.

\bibitem[{Benz and Jasinskaja(2017)}]{benz2017questions}
Anton Benz and Katja Jasinskaja. 2017.
\newblock Questions under discussion: From sentence to discourse.
\newblock \emph{Discourse Processes}, 54(3):177--186.

\bibitem[{Bhatia et~al.(2015)Bhatia, Ji, and
  Eisenstein}]{bhatia-etal-2015-better}
Parminder Bhatia, Yangfeng Ji, and Jacob Eisenstein. 2015.
\newblock Better document-level sentiment analysis from {RST} discourse
  parsing.
\newblock In \emph{Proceedings of the 2015 Conference on Empirical Methods in
  Natural Language Processing}, pages 2212--2218.

\bibitem[{Bott and Solstad(2014)}]{bott2014verbs}
Oliver Bott and Torgrim Solstad. 2014.
\newblock From verbs to discourse: A novel account of implicit causality.
\newblock In \emph{Psycholinguistic approaches to meaning and understanding
  across languages}, pages 213--251.

\bibitem[{B{\"u}ring(2003)}]{buring2003d}
Daniel B{\"u}ring. 2003.
\newblock On d-trees, beans, and b-accents.
\newblock \emph{Linguistics and philosophy}, 26(5):511--545.

\bibitem[{Carlson and Marcu(2001)}]{carlson2001discourse}
Lynn Carlson and Daniel Marcu. 2001.
\newblock Discourse tagging reference manual.
\newblock \emph{ISI Technical Report ISI-TR-545}, 54(2001):56.

\bibitem[{Carlson et~al.(2001)Carlson, Marcu, and Okurovsky}]{rstdt}
Lynn Carlson, Daniel Marcu, and Mary~Ellen Okurovsky. 2001.
\newblock Building a discourse-tagged corpus in the framework of rhetorical
  structure theory.
\newblock In \emph{{ SIGdial Workshop on Discourse and Dialogue}}.

\bibitem[{Celikyilmaz et~al.(2020)Celikyilmaz, Clark, and
  Gao}]{celikyilmaz2020evaluation}
Asli Celikyilmaz, Elizabeth Clark, and Jianfeng Gao. 2020.
\newblock Evaluation of text generation: A survey.
\newblock \emph{arXiv preprint arXiv:2006.14799}.

\bibitem[{De~Kuthy et~al.(2018)De~Kuthy, Reiter, and Riester}]{de2018qud}
Kordula De~Kuthy, Nils Reiter, and Arndt Riester. 2018.
\newblock Qud-based annotation of discourse structure and information
  structure: Tool and evaluation.
\newblock In \emph{Proceedings of the Eleventh International Conference on
  Language Resources and Evaluation}.

\bibitem[{Demberg et~al.(2019)Demberg, Scholman, and
  Asr}]{demberg2019compatible}
Vera Demberg, Merel~CJ Scholman, and Fatemeh~Torabi Asr. 2019.
\newblock How compatible are our discourse annotation frameworks? insights from
  mapping rst-dt and pdtb annotations.
\newblock \emph{Dialogue \& Discourse}, 10(1):87--135.

\bibitem[{Devlin et~al.(2019)Devlin, Chang, Lee, and Toutanova}]{BERT}
Jacob Devlin, Ming-Wei Chang, Kenton Lee, and Kristina Toutanova. 2019.
\newblock {BERT}: Pre-training of deep bidirectional transformers for language
  understanding.
\newblock In \emph{Proceedings of the 2019 Conference of the North American
  Chapter of the Association for Computational Linguistics: Human Language
  Technologies, Volume 1 (Long and Short Papers)}, pages 4171--4186.

\bibitem[{Durrett et~al.(2016)Durrett, Berg-Kirkpatrick, and
  Klein}]{durrett2016learning}
Greg Durrett, Taylor Berg-Kirkpatrick, and Dan Klein. 2016.
\newblock Learning-based single-document summarization with compression and
  anaphoricity constraints.
\newblock In \emph{Proceedings of the 54th Annual Meeting of the Association
  for Computational Linguistics (Volume 1: Long Papers)}, pages 1998--2008.

\bibitem[{Ferracane et~al.(2019)Ferracane, Durrett, Li, and
  Erk}]{ferracane-etal-2019-evaluating}
Elisa Ferracane, Greg Durrett, Junyi~Jessy Li, and Katrin Erk. 2019.
\newblock Evaluating discourse in structured text representations.
\newblock In \emph{Proceedings of the 57th Annual Meeting of the Association
  for Computational Linguistics}, pages 646--653.

\bibitem[{Garvey and Caramazza(1974)}]{garvey1974implicit}
Catherine Garvey and Alfonso Caramazza. 1974.
\newblock Implicit causality in verbs.
\newblock \emph{Linguistic inquiry}, 5(3):459--464.

\bibitem[{Gerani et~al.(2014)Gerani, Mehdad, Carenini, Ng, and
  Nejat}]{gerani-etal-2014-abstractive}
Shima Gerani, Yashar Mehdad, Giuseppe Carenini, Raymond Ng, and Bita Nejat.
  2014.
\newblock Abstractive summarization of product reviews using discourse
  structure.
\newblock In \emph{Proceedings of the 2014 conference on empirical methods in
  natural language processing}, pages 1602--1613.

\bibitem[{Ginzburg et~al.(1996)}]{ginzburg1996dynamics}
Jonathan Ginzburg et~al. 1996.
\newblock Dynamics and the semantics of dialogue.
\newblock \emph{Logic, language and computation}, 1:221--237.

\bibitem[{Grice(1975)}]{grice1975logic}
Herbert~P Grice. 1975.
\newblock Logic and conversation.
\newblock In \emph{Speech acts}, pages 41--58. Brill.

\bibitem[{He et~al.(2015)He, Lewis, and Zettlemoyer}]{he2015question}
Luheng He, Mike Lewis, and Luke Zettlemoyer. 2015.
\newblock Question-answer driven semantic role labeling: Using natural language
  to annotate natural language.
\newblock In \emph{Proceedings of the 2015 conference on empirical methods in
  natural language processing}, pages 643--653.

\bibitem[{Hesse et~al.(2020)Hesse, Benz, Langner, Theodor, and
  Klabunde}]{hesse2020annotating}
Christoph Hesse, Anton Benz, Maurice Langner, Felix Theodor, and Ralf Klabunde.
  2020.
\newblock Annotating quds for generating pragmatically rich texts.
\newblock In \emph{Proceedings of the Workshop on Discourse Theories for Text
  Planning}, pages 10--16.

\bibitem[{Hirao et~al.(2013)Hirao, Yoshida, Nishino, Yasuda, and
  Nagata}]{hirao2013single}
Tsutomu Hirao, Yasuhisa Yoshida, Masaaki Nishino, Norihito Yasuda, and Masaaki
  Nagata. 2013.
\newblock Single-document summarization as a tree knapsack problem.
\newblock In \emph{Proceedings of the 2013 conference on empirical methods in
  natural language processing}, pages 1515--1520.

\bibitem[{Hirschberg(1985)}]{hirschberg1985theory}
Julia Linn~Bell Hirschberg. 1985.
\newblock \emph{A theory of scalar implicature}.
\newblock University of Pennsylvania.

\bibitem[{Holtzman et~al.(2020)Holtzman, Buys, Du, Forbes, and Choi}]{nucleus}
Ari Holtzman, Jan Buys, Li~Du, Maxwell Forbes, and Yejin Choi. 2020.
\newblock {The Curious Case of Neural Text Degeneration }.
\newblock In \emph{International Conference on Learning Representations}.

\bibitem[{Howcroft et~al.(2020)Howcroft, Belz, Clinciu, Gkatzia, Hasan,
  Mahamood, Mille, Van~Miltenburg, Santhanam, and Rieser}]{howcroft2020twenty}
David~M Howcroft, Anja Belz, Miruna-Adriana Clinciu, Dimitra Gkatzia, Sadid~A
  Hasan, Saad Mahamood, Simon Mille, Emiel Van~Miltenburg, Sashank Santhanam,
  and Verena Rieser. 2020.
\newblock Twenty years of confusion in human evaluation: Nlg needs evaluation
  sheets and standardised definitions.
\newblock In \emph{Proceedings of the 13th International Conference on Natural
  Language Generation}, pages 169--182.

\bibitem[{Hunter and Abrus{\'a}n(2015)}]{hunter2015rhetorical}
Julie Hunter and M{\'a}rta Abrus{\'a}n. 2015.
\newblock Rhetorical structure and quds.
\newblock In \emph{JSAI International Symposium on Artificial Intelligence},
  pages 41--57.

\bibitem[{Jasinskaja et~al.(2017)Jasinskaja, Salfner, and
  Freitag}]{jasinskaja2017discourse}
Katja Jasinskaja, Fabienne Salfner, and Constantin Freitag. 2017.
\newblock Discourse-level implicature: A case for qud.
\newblock \emph{Discourse Processes}, 54(3):239--258.

\bibitem[{Jasinskaja et~al.(2008)Jasinskaja, Zeevat
  et~al.}]{jasinskaja2008explaining}
Katja Jasinskaja, Henk Zeevat, et~al. 2008.
\newblock Explaining additive, adversative and contrast marking in russian and
  english.
\newblock \emph{Revue de S{\'e}mantique et Pragmatique}, 24(1):65--91.

\bibitem[{Ji and Smith(2017)}]{ji2017neural}
Yangfeng Ji and Noah~A Smith. 2017.
\newblock Neural discourse structure for text categorization.
\newblock In \emph{Proceedings of the 55th Annual Meeting of the Association
  for Computational Linguistics (Volume 1: Long Papers)}, pages 996--1005.

\bibitem[{Kehler et~al.(2008)Kehler, Kertz, Rohde, and
  Elman}]{kehler2008coherence}
Andrew Kehler, Laura Kertz, Hannah Rohde, and Jeffrey~L Elman. 2008.
\newblock Coherence and coreference revisited.
\newblock \emph{Journal of semantics}, 25(1):1--44.

\bibitem[{Kehler and Rohde(2017)}]{kehler2017evaluating}
Andrew Kehler and Hannah Rohde. 2017.
\newblock Evaluating an expectation-driven question-under-discussion model of
  discourse interpretation.
\newblock \emph{Discourse Processes}, 54(3):219--238.

\bibitem[{Kingma and Ba(2015)}]{adam}
Diederik~P. Kingma and Jimmy Ba. 2015.
\newblock Adam: A method for stochastic optimization.
\newblock In \emph{Proceedings of the 3rd International Conference for Learning
  Representations}.

\bibitem[{Klein et~al.(2020)Klein, Mamou, Pyatkin, Stepanov, He, Roth,
  Zettlemoyer, and Dagan}]{klein2020qanom}
Ayal Klein, Jonathan Mamou, Valentina Pyatkin, Daniela Stepanov, Hangfeng He,
  Dan Roth, Luke Zettlemoyer, and Ido Dagan. 2020.
\newblock Qanom: Question-answer driven srl for nominalizations.
\newblock In \emph{Proceedings of the 28th International Conference on
  Computational Linguistics}, pages 3069--3083.

\bibitem[{Ko et~al.(2020)Ko, Chen, Huang, Durrett, and Li}]{inq}
Wei-Jen Ko, Te-yuan Chen, Yiyan Huang, Greg Durrett, and Junyi~Jessy Li. 2020.
\newblock Inquisitive question generation for high level text comprehension.
\newblock In \emph{Proceedings of the 2020 Conference on Empirical Methods in
  Natural Language Processing}, pages 6544--6555.

\bibitem[{Ko et~al.(2022)Ko, Dalton, Simmons, Fisher, Durrett, and Li}]{dcqa}
Wei-Jen Ko, Cutter Dalton, Mark Simmons, Eliza Fisher, Greg Durrett, and
  Junyi~Jessy Li. 2022.
\newblock Discourse comprehension: A question answering framework to represent
  sentence connections.
\newblock In \emph{Proceedings of the 2022 Conference on Empirical Methods in
  Natural Language Processing}, pages 11752--11764.

\bibitem[{Kuppevelt(1996)}]{kuppevelt1996directionality}
Jan~van Kuppevelt. 1996.
\newblock Directionality in discourse: Prominence differences in subordination
  relations1.
\newblock \emph{Journal of semantics}, 13(4):363--395.

\bibitem[{Lee and Goldwasser(2019)}]{lee2019multi}
I-Ta Lee and Dan Goldwasser. 2019.
\newblock Multi-relational script learning for discourse relations.
\newblock In \emph{Proceedings of the 57th Annual Meeting of the Association
  for Computational Linguistics}, pages 4214--4226.

\bibitem[{Mann and Thompson(1988)}]{rst}
William~C Mann and Sandra~A Thompson. 1988.
\newblock Rhetorical structure theory: Toward a functional theory of text
  organization.
\newblock \emph{Text-interdisciplinary Journal for the Study of Discourse},
  8(3):243--281.

\bibitem[{Marcu(1996)}]{marcu1996building}
Daniel Marcu. 1996.
\newblock Building up rhetorical structure trees.
\newblock In \emph{Proceedings of the National Conference on Artificial
  Intelligence}, pages 1069--1074.

\bibitem[{Marcu(1998)}]{marcu1998build}
Daniel Marcu. 1998.
\newblock To build text summaries of high quality, nuclearity is not
  sufficient.
\newblock In \emph{Working Notes of the AAAI-98 Spring Symposium on Intelligent
  Text Summarization}, pages 1--8.

\bibitem[{Morey et~al.(2018)Morey, Muller, and Asher}]{morey2018dependency}
Mathieu Morey, Philippe Muller, and Nicholas Asher. 2018.
\newblock A dependency perspective on rst discourse parsing and evaluation.
\newblock \emph{Computational Linguistics}, 44(2):197--235.

\bibitem[{Narasimhan and Barzilay(2015)}]{narasimhan2015machine}
Karthik Narasimhan and Regina Barzilay. 2015.
\newblock Machine comprehension with discourse relations.
\newblock In \emph{Proceedings of the 53rd Annual Meeting of the Association
  for Computational Linguistics and the 7th International Joint Conference on
  Natural Language Processing (Volume 1: Long Papers)}, pages 1253--1262.

\bibitem[{Onea(2016)}]{onea2016potential}
Edgar Onea. 2016.
\newblock \emph{Potential questions at the semantics-pragmatics interface}.
\newblock Brill.

\bibitem[{Petersen and Ostendorf(2007)}]{petersen2007text}
Sarah~E Petersen and Mari Ostendorf. 2007.
\newblock Text simplification for language learners: a corpus analysis.
\newblock In \emph{Workshop on speech and language technology in education}.

\bibitem[{Prasad et~al.(2008)Prasad, Dinesh, Lee, Miltsakaki, Robaldo, Joshi,
  and Webber}]{pdtb}
Rashmi Prasad, Nikhil Dinesh, Alan Lee, Eleni Miltsakaki, Livio Robaldo,
  Aravind~K Joshi, and Bonnie~L Webber. 2008.
\newblock {The Penn Discourse TreeBank 2.0.}
\newblock In \emph{Language Resources and Evaluation Conference}.

\bibitem[{Prasad et~al.(2014)Prasad, Webber, and Joshi}]{prasad2014reflections}
Rashmi Prasad, Bonnie Webber, and Aravind Joshi. 2014.
\newblock Reflections on the penn discourse treebank, comparable corpora, and
  complementary annotation.
\newblock \emph{Computational Linguistics}, 40(4):921--950.

\bibitem[{Pyatkin et~al.(2020)Pyatkin, Klein, Tsarfaty, and
  Dagan}]{pyatkin2020qadiscourse}
Valentina Pyatkin, Ayal Klein, Reut Tsarfaty, and Ido Dagan. 2020.
\newblock {QADiscourse}-discourse relations as qa pairs: Representation,
  crowdsourcing and baselines.
\newblock In \emph{Proceedings of the 2020 Conference on Empirical Methods in
  Natural Language Processing}, pages 2804--2819.

\bibitem[{Radford et~al.(2019)Radford, Wu, Child, Luan, Amodei, and
  Sutskever}]{gpt2}
Alec Radford, Jeffrey Wu, Rewon Child, David Luan, Dario Amodei, and Ilya
  Sutskever. 2019.
\newblock Language models are unsupervised multitask learners.
\newblock \emph{OpenAI Technical Report}.

\bibitem[{Riester(2019)}]{riester2019constructing}
Arndt Riester. 2019.
\newblock Constructing {QUD} trees.
\newblock In \emph{Questions in discourse}, pages 164--193. Brill.

\bibitem[{Roberts(2004)}]{roberts2004context}
Craige Roberts. 2004.
\newblock Context in dynamic interpretation.
\newblock \emph{The handbook of pragmatics}, 197:220.

\bibitem[{Roberts(2012)}]{roberts2012information}
Craige Roberts. 2012.
\newblock Information structure: Towards an integrated formal theory of
  pragmatics.
\newblock \emph{Semantics and pragmatics}, 5:6--1.

\bibitem[{Sanders et~al.(1992)Sanders, Spooren, and
  Noordman}]{sanders1992toward}
Ted~JM Sanders, Wilbert~PM Spooren, and Leo~GM Noordman. 1992.
\newblock Toward a taxonomy of coherence relations.
\newblock \emph{Discourse processes}, 15(1):1--35.

\bibitem[{Sang et~al.(2003)Sang, F., and
  De~Meulder}]{tjong-kim-sang-de-meulder-2003-introduction}
Tjong~Kim Sang, Erik F., and Fien De~Meulder. 2003.
\newblock Introduction to the {C}o{NLL}-2003 shared task: Language-independent
  named entity recognition.
\newblock In \emph{Proceedings of the Seventh Conference on Natural Language
  Learning at {HLT}-{NAACL} 2003}.

\bibitem[{Simons et~al.(2010)Simons, Tonhauser, Beaver, and
  Roberts}]{simons2010projects}
Mandy Simons, Judith Tonhauser, David Beaver, and Craige Roberts. 2010.
\newblock What projects and why.
\newblock In \emph{Semantics and linguistic theory}, volume~20, pages 309--327.

\bibitem[{Srikanth and Li(2021)}]{srikanth2021elaborative}
Neha Srikanth and Junyi~Jessy Li. 2021.
\newblock Elaborative simplification: Content addition and explanation
  generation in text simplification.
\newblock In \emph{Findings of the Association for Computational Linguistics:
  ACL-IJCNLP 2021}, pages 5123--5137.

\bibitem[{Stalnaker(1978)}]{stalnaker1978assertion}
Robert~C Stalnaker. 1978.
\newblock Assertion.
\newblock In \emph{Pragmatics}, pages 315--332. Brill.

\bibitem[{Taboada and Mann(2006)}]{taboada2006rhetorical}
Maite Taboada and William~C Mann. 2006.
\newblock Rhetorical structure theory: Looking back and moving ahead.
\newblock \emph{Discourse studies}, 8(3):423--459.

\bibitem[{Van~Kuppevelt(1995)}]{van1995discourse}
Jan Van~Kuppevelt. 1995.
\newblock Discourse structure, topicality and questioning.
\newblock \emph{Journal of linguistics}, 31(1):109--147.

\bibitem[{Van~Kuppevelt(1996)}]{van1996inferring}
Jan Van~Kuppevelt. 1996.
\newblock Inferring from topics: Scalar implicatures as topic-dependent
  inferences.
\newblock \emph{Linguistics and philosophy}, pages 393--443.

\bibitem[{Velleman and Beaver(2016)}]{velleman2016question}
Leah Velleman and David Beaver. 2016.
\newblock Question-based models of information structure.
\newblock In \emph{The Oxford handbook of information structure}.

\bibitem[{Von~Stutterheim and Klein(1989)}]{von1989referential}
Christiane Von~Stutterheim and Wolfgang Klein. 1989.
\newblock Referential movement in descriptive and narrative discourse.
\newblock In \emph{North-Holland Linguistic Series: Linguistic Variations},
  volume~54, pages 39--76.

\bibitem[{Westera et~al.(2020)Westera, Mayol, and Rohde}]{tedq}
Matthijs Westera, Laia Mayol, and Hannah Rohde. 2020.
\newblock {TED}-{Q}: {TED} talks and the questions they evoke.
\newblock In \emph{Proceedings of The 12th Language Resources and Evaluation
  Conference}, pages 1118--1127.

\bibitem[{Wolf and Gibson(2005)}]{wolf2005representing}
Florian Wolf and Edward Gibson. 2005.
\newblock Representing discourse coherence: A corpus-based study.
\newblock \emph{Computational linguistics}, 31(2):249--287.

\bibitem[{Wolf et~al.(2020)Wolf, Debut, Sanh, Chaumond, Delangue, Moi, Cistac,
  Rault, Louf, Funtowicz, Davison, Shleifer, von Platen, Ma, Jernite, Plu, Xu,
  Scao, Gugger, Drame, Lhoest, and Rush}]{wolf-etal-2020-transformers}
Thomas Wolf, Lysandre Debut, Victor Sanh, Julien Chaumond, Clement Delangue,
  Anthony Moi, Pierric Cistac, Tim Rault, Rémi Louf, Morgan Funtowicz, Joe
  Davison, Sam Shleifer, Patrick von Platen, Clara Ma, Yacine Jernite, Julien
  Plu, Canwen Xu, Teven~Le Scao, Sylvain Gugger, Mariama Drame, Quentin Lhoest,
  and Alexander~M. Rush. 2020.
\newblock Transformers: State-of-the-art natural language processing.
\newblock In \emph{Proceedings of the 2020 Conference on Empirical Methods in
  Natural Language Processing: System Demonstrations}, pages 38--45.

\bibitem[{Xu et~al.(2020)Xu, Gan, Cheng, and Liu}]{xu2020discourse}
Jiacheng Xu, Zhe Gan, Yu~Cheng, and Jingjing Liu. 2020.
\newblock Discourse-aware neural extractive text summarization.
\newblock In \emph{Proceedings of the 58th Annual Meeting of the Association
  for Computational Linguistics}, pages 5021--5031.

\bibitem[{Xu et~al.(2015)Xu, Callison-Burch, and Napoles}]{newsela}
Wei Xu, Chris Callison-Burch, and Courtney Napoles. 2015.
\newblock Problems in current text simplification research: New data can help.
\newblock \emph{Transactions of the Association for Computational Linguistics},
  3:283--297.

\bibitem[{Yadav et~al.(2019)Yadav, Husain, and
  Futrell}]{yadav-etal-2019-formal}
Himanshu Yadav, Samar Husain, and Richard Futrell. 2019.
\newblock Are formal restrictions on crossing dependencies epiphenominal?
\newblock In \emph{Proceedings of the 18th International Workshop on Treebanks
  and Linguistic Theories (TLT, SyntaxFest 2019)}, pages 2--12.

\bibitem[{Zhong et~al.(2020)Zhong, Jiang, Xu, and Li}]{zhong2020discourse}
Yang Zhong, Chao Jiang, Wei Xu, and Junyi~Jessy Li. 2020.
\newblock Discourse level factors for sentence deletion in text simplification.
\newblock In \emph{Proceedings of the AAAI Conference on Artificial
  Intelligence}, volume~34, pages 9709--9716.

\bibitem[{Zhu et~al.(2010)Zhu, Bernhard, and Gurevych}]{zhu2010monolingual}
Zhemin Zhu, Delphine Bernhard, and Iryna Gurevych. 2010.
\newblock A monolingual tree-based translation model for sentence
  simplification.
\newblock In \emph{Proceedings of the 23rd International Conference on
  Computational Linguistics}, pages 1353--1361.

\end{thebibliography}
\bibliographystyle{acl_natbib}

\appendix

\section{Inter-annotator agreement for human judgments}
\label{app:agreement}
For \textbf{Question 1}, the three annotators all agree on 54\% of the fine grain labels, and there is a majority on 93\%  of questions on fine grained labels. 
Krippendorff's alpha is 0.366 for ``yes'' vs.\ others, 0.319 for the 4 coarse categories, and 0.317 for all labels at the most fine-grained label. 
For \textbf{Question 2}, the three annotators all agree on 60\% of the fine grain labels, and there is a majority on 93\%  of questions on fine grained labels. Krippendorff's alpha is 0.376 for ``yes'' vs.\ others, and 0.297 for the 4 categories.

All the alpha values above indicate ``fair'' agreement~\cite{artstein2008inter}. One reason for this is a clear majority of ``yes'' labels for both questions; nonetheless these values indicate a certain degree of subjectivity in the tasks. 

\section{Reranker details}\label{app:reranker}
To train the reranker, we use questions in the DCQA training set as positive examples, and swap the answer or the anchor sentence with \emph{every} other sentence from the same article to create negative examples.
This resulted in a training set of 709,532 instances.
We fine-tune the BERT model on this data for 3 epochs using learning rate 2e-5 and batch size 32, trained using binary cross entropy loss. 

On the DCQA validation set, among about 37 options generated from the same question, the ranks of the correct response predicted by the model is on the 14\% percentile in average.

\section{Anchor prediction}\label{app:anchor_prediction}
We also report the accuracy of the predicted anchor sentences for the first part of our pipeline model (i.e., before the questions get generated). Note that this is a partial notion of accuracy for analysis purposes, since it is natural for different questions to be triggered form different sentences (and sometimes perfectly fine for the same question to come from different sentences)~\cite{dcqa}.
On the validation and test set of the DCQA dataset, and the agreement between the model and human on 46.8\% of the instances (the annotations of different annotators are treated as separate instances).
This is the same as DCQA's statistics between two human annotators.

\section{Example model outputs}\label{app:outputs}

We show an additional snippet of example model output:

\begin{quote}
    \small 
    \textbf{Context: [9]} In 1971, Sierra Nevada bighorns were one of the first animals listed as threatened under the California Endangered Species Act. \textbf{[10]} In 2000, the federal government added the bighorns to its endangered lists. \textbf{[11]} ‘There was a lot of concern about extinction,’ says state biologist Tom Stephenson, the recovery project leader. \textbf{[12]} ‘But with some good fortune and the combination of the right recovery efforts, it’s gone as well as anybody could’ve imagined’. \textbf{[13]} Teams of biologists and volunteers in 2000 began their research, and in 2007 started reintroducing the Sierra Nevada bighorn by dispersing them into herds along the Sierra's crest. \textbf{[14]} The agencies designated 16 areas for the bighorns with the initial goal of repopulating 12 of them.
    
    \textbf{9-10}: What happened after that?
    
    \textbf{10-11}: What was the opinion of those involved in the recovery project?
    
    \textbf{9-12}: What happened to the bighorns?
    
    \textbf{12-13}: How did recovery efforts eventually go?
    
    \textbf{13-14}: How many areas were to be re-population based on the initial work?

\end{quote}

\section{Compute}
For all models in this work, we used 2 compute nodes each consisting of 3x NVIDIA A100 GPUs.

\end{document}